%% file: main.tex
\pdfoutput=1

\documentclass[11pt]{article}

\usepackage{emnlp2021}
\usepackage{times}
\usepackage{latexsym}

\usepackage[T1]{fontenc}

\usepackage[utf8]{inputenc}

\usepackage{microtype}

\usepackage{microtype}
\usepackage{epsfig}
\usepackage{multirow}
\usepackage{amsmath}
\usepackage{amsthm}
\usepackage{graphicx}
\usepackage{amssymb}
\usepackage{subcaption}
\usepackage{enumitem}
\usepackage{caption}
\usepackage{arydshln}
\usepackage[T1]{fontenc}
\usepackage[utf8]{inputenc}

\author{\stepcounter{footnote}Siamak Shakeri \thanks{\ \ Corresponding author.}\\
  Google Research \\
  \texttt{siamaks@google.com} \\\And
    Noah Constant \\
  Google Research \\
  \texttt{nconstant@google.com} \\\AND
  Mihir Sanjay Kale \\
  Google Research \\
  \texttt{mihirkale@google.com} \\\And
  Linting Xue \\
  Google Research \\
  \texttt{lintingx@google.com} 
  }

\date{}

\begin{document}

\newcommand\BibTeX{B\textsc{ib}\TeX}
\renewcommand{\thefootnote}{\fnsymbol{footnote}}

\title{Towards Zero-Shot Multilingual Synthetic Question and Answer Generation for
Cross-Lingual Reading Comprehension}

\maketitle

\begin{abstract}
We propose a simple method to generate multilingual question and answer pairs on a large scale through the use of a single generative model. These synthetic samples can be used to improve the zero-shot performance of multilingual QA models on target languages. Our proposed multi-task training of the generative model only requires the labeled training samples in English, thus removing the need for such samples in the target languages, making it applicable to far more languages than those with labeled data. Human evaluations indicate the majority of such samples are grammatically correct and sensible. Experimental results show our proposed approach can achieve large gains on the XQuAD dataset, reducing the gap between zero-shot and supervised performance of smaller QA models on various languages.

\end{abstract}
\input{introduction}
\input{method.tex}
\input{related_work.tex}
\input{experiments.tex}
\input{conclusion.tex}
\input{acknowledgement}
\newpage
\clearpage
\bibliography{anthology,custom}
\bibliographystyle{acl_natbib}

\end{document}

%% file: introduction.tex
\section{Introduction}
Generating question and answers from raw text has always been a challenging problem in natural language generation. Recently, there have been numerous efforts around question generation \cite{du2017learning,song-etal-2018-leveraging,klein2019learning,wangneural,ma2020improving,Chen2020Reinforcement,tuan2020capturing}.

Using such synthetic samples to improve the performance of question answering models has been explored by
\citet{puri2020nvidia}, \citet{alberti-etal-2019-synthetic}, and \citet{shakeri2020endtoend}, who show that reading comprehension (RC) models can be improved by generating large-scale synthetic training data. These promising results combined with the recent surge in the development of powerful generative models such as GPT-3 \cite{gpt3}, BART \cite{lewis2019bart}, and T5 \cite{raffel2019t5} suggest that the need for large manually labeled datasets can be reduced.

Although synthetic question-answer (QA) generation is well explored in English, the efficacy of such methods in the other languages remains an open question. Considering the lack of manually labeled QA datasets in many languages other than English, QA generation techniques can be applied to improve RC models in those languages. The emergence of multilingual generative models such as mBART \cite{mbart} and mT5 \cite{mt5} facilitates such endeavors.

\begin{figure}
    \centering
    \includegraphics[width=\columnwidth]{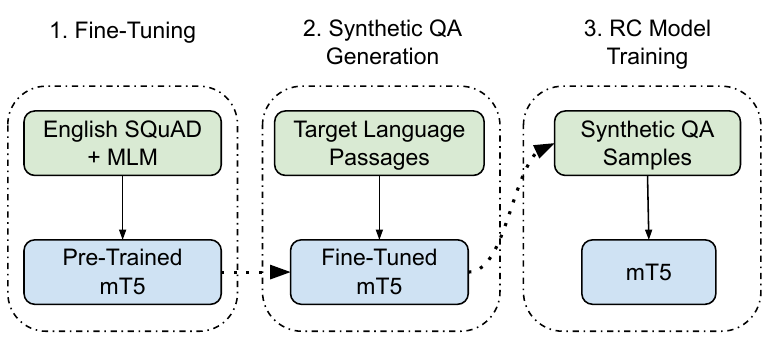}
    \caption{\small{End-to-End pipeline: 1)~Fine-tuning the generative model using SQuAD English samples and multilingual MLM\@. 2)~Generating synthetic samples from Wikipedia passages of the target language using the fine-tuned generative model. 3)~Training the downstream reading comprehension model using synthetic samples.}}
  \label{fig:pipeline}
\end{figure}

In this work, we propose generating multilingual question answer pairs to improve the performance of RC models in languages other than English. Besides unlabeled articles and questions, Our proposed method only requires labeled training samples in English, thus completely removing the need to acquire new labeled datasets. Our approach can easily be extended to any language, as long the multilingual generative model supports the language, and unlabeled questions and articles, such as Wikipedia, books, etc., exist in that language.

To enable zero-shot QA generation, the generative model should be able to produce non-English QA samples on non-English input when only trained on English samples. Inspired by the work of \citet{artetxe-etal-2020-cross,gururangan-etal-2020-dont,liu2020exploring}, we propose a multi-task learning setting, where during the fine-tuning stage, we train on two tasks in parallel: the target question-answer generation task, and the multilingual masked language modeling (MLM) task that was used in pre-training the generative model. Our experimental results show that including the MLM task is crucial in enabling the zero-shot capability of the fine-tuned generative model.

We propose fine-tuning a pre-trained multilingual T5 model on the SQuAD 1.1 \cite{squad11} training set. The fine-tuned model is then used to generate a large set of synthetic question-answer pairs from Wikipedia passages in the target language. Fig.~\ref{fig:pipeline} illustrates the end-to-end pipeline. We show that such synthetic samples can significantly boost RC models trained only on the English samples, with improvements up to 9 absolute points on F1\@. To summarize, our contributions are:

\begin{itemize}[topsep=3pt,itemsep=0ex,partopsep=1ex,parsep=1ex] 
    \item  Improving the zero-shot performance of multilingual RC models on multilingual QA tasks through generation of synthetic multilingual QA pairs.
    \item Proposing a multi-task fine-tuning of the multilingual generative model which is crucial in enabling zero-shot multilingual generation.
    \item Our approach is entirely zero-shot. No manually-labeled sample is used in fine-tuning of generative models on target languages, applicable to both high and low resource languages.
    \item Demonstrating grammatical correctness and sensibility of generated questions through human evaluations.
\end{itemize}

The rest of the paper is organized as follows. In section \ref{sec:e2e}, we discuss the process designed to train the generative model and produce synthetic samples. Section \ref{sec:related} discusses related work in the area of multilingual question-answer generation. In section \ref{sec:experiments}, we present experiments to measure the quality of generated samples. Section \ref{sec:downstream_experiments} focuses on the application of synthetic question-answer samples to downstream reading comprehension models. Finally, we conclude in section \ref{sec:conclusion}.

%% file: method.tex
\section{End-to-End Question-Answer Generation and Filtering}\label{sec:e2e}

\subsection{Modeling}\label{sec:modeling}
We use pre-trained ``multilingual T5'' (mT5) \cite{mt5} as our generative model. The mT5 model is based on T5 \cite{raffel2019t5}, which is an encoder-decoder sequence-to-sequence model.
\subsection{QA Generation Task}\label{sec:qa_gen_task}
We follow the probability distribution factorization suggested by \citet{shakeri2020endtoend}, where:
\begin{align*}
  p(Q,A | P) = p(Q | P) \times p(A | Q, P)
\end{align*}
Sampling from the above factorization is performed as follows:
\begin{align*}
    q \sim p(Q|P), a \sim p(A|Q,P) 
\end{align*}
where $Q,P,A$ refer to question, passage, and answer, respectively. 
During fine-tuning, passage tokens are fed as inputs, and the targets are a concatenation of the question and answer tokens.
During sampling, candidate passages are passed as inputs to the fine-tuned generative model, and question-answer pairs are sampled from the decoder.

\begin{figure*}[h!]
    \centering
    \includegraphics[width=\textwidth]{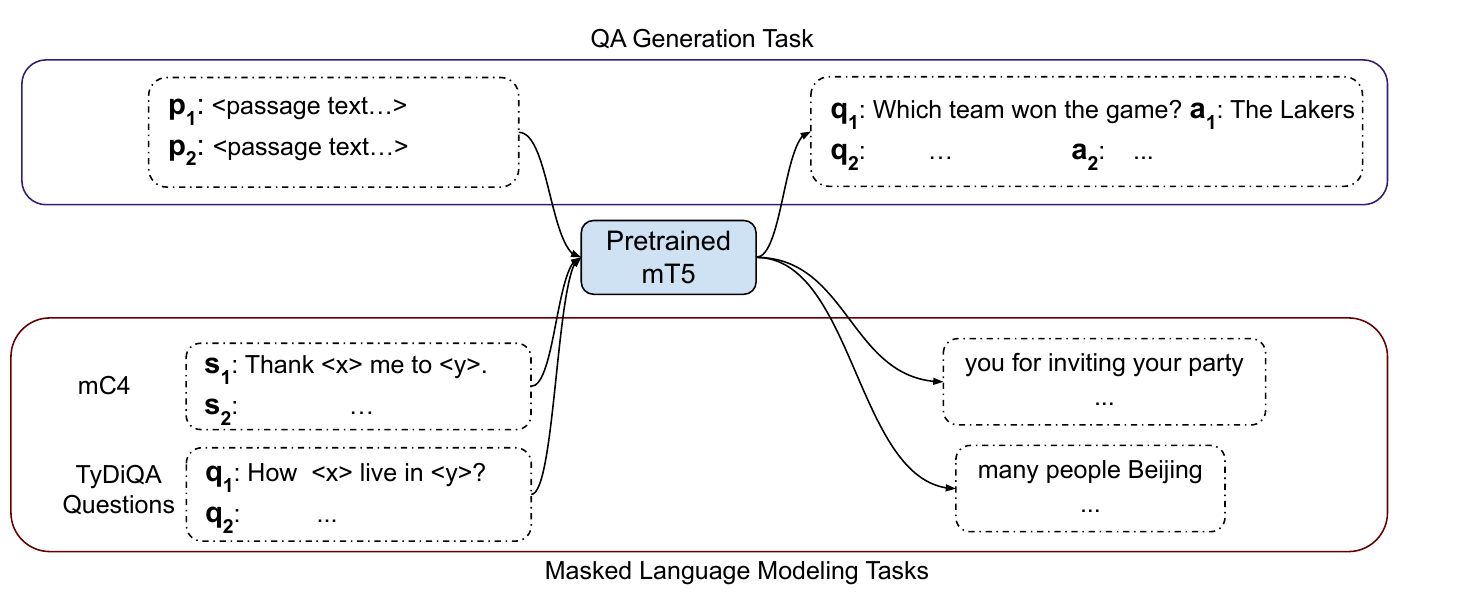}
    \caption{\small{Multi-task fine-tuning of the multilingual pre-trained mT5 model: 1)~QA generation task, which uses SQuAD English samples, 2)~MLM task on a subset of mC4, 3)~MLM on the questions from the TyDiQA Gold Passage Task. The MLM variant used does not include sentinel tokens in the decoder output.}}
  \label{fig:model}
\end{figure*}

Fig.~\ref{fig:model} depicts the fine-tuning and sampling processes. We prepend \textit{``question''} to the question tokens and \textit{``answer''} to the answer tokens, to help the model distinguish one from the other.
\subsection{Masked Language Modeling Task}\label{sec:mlm}
The mT5 model is pre-trained on the large multilingual ``mC4'' dataset \cite{mt5} built from Common Crawl data, and trained using a Masked Language Modeling (MLM) task. This task involves replacing contiguous spans of input tokens with unique sentinel tokens (one per span). The decoder is then trained to reconstruct all the masked spans in the input, using a standard cross-entropy loss. We found a variant of this MLM task, where we remove all ``sentinel'' tokens (corresponding to non-masked spans in the input text) from the target sequence, as we notice this improves the quality of generated QAs. This variant is used in this work.

\subsection{Multi-Task Fine-Tuning}\label{sec:multi-task}
To perform zero-shot generation, the model needs to not only learn the QA Generation task, but also to retain its multilingual generation capabilities achieved during pre-training. To avoid \textit{catastrophic forgetting} \cite{FRENCH1999128}, which could lead to degraded generation capability, we propose a multi-task setting, where a predetermined percentage of fine-tuning examples comes from the QA Generation task, while the remaining examples (trained in parallel) are from a mixture of two MLM tasks: 1) MLM on a subset of mC4 which is a continuation of mT5 pre-training, 2) MLM on the questions from TyDiQA Gold Passage dev and training sets. The MLM on mC4 task helps the fine-tuned model retain its multi-lingual generation capabilities, while the MLM on TyDiQA questions task further improves the question generation capabilities of the generative model. Please note that the only supervised QA training data is SQuAD 1.1. The MLM task on TyDiQA questions is not conditioned on the associated passages of the questions. Experimental results in section \ref{sec:experiments} demonstrate the efficacy of this proposed approach. Fig.~\ref{fig:model} illustrates the multi-task fine-tuning process.

\begin{figure*}[h]
    \centering
    \includegraphics[scale=0.25]{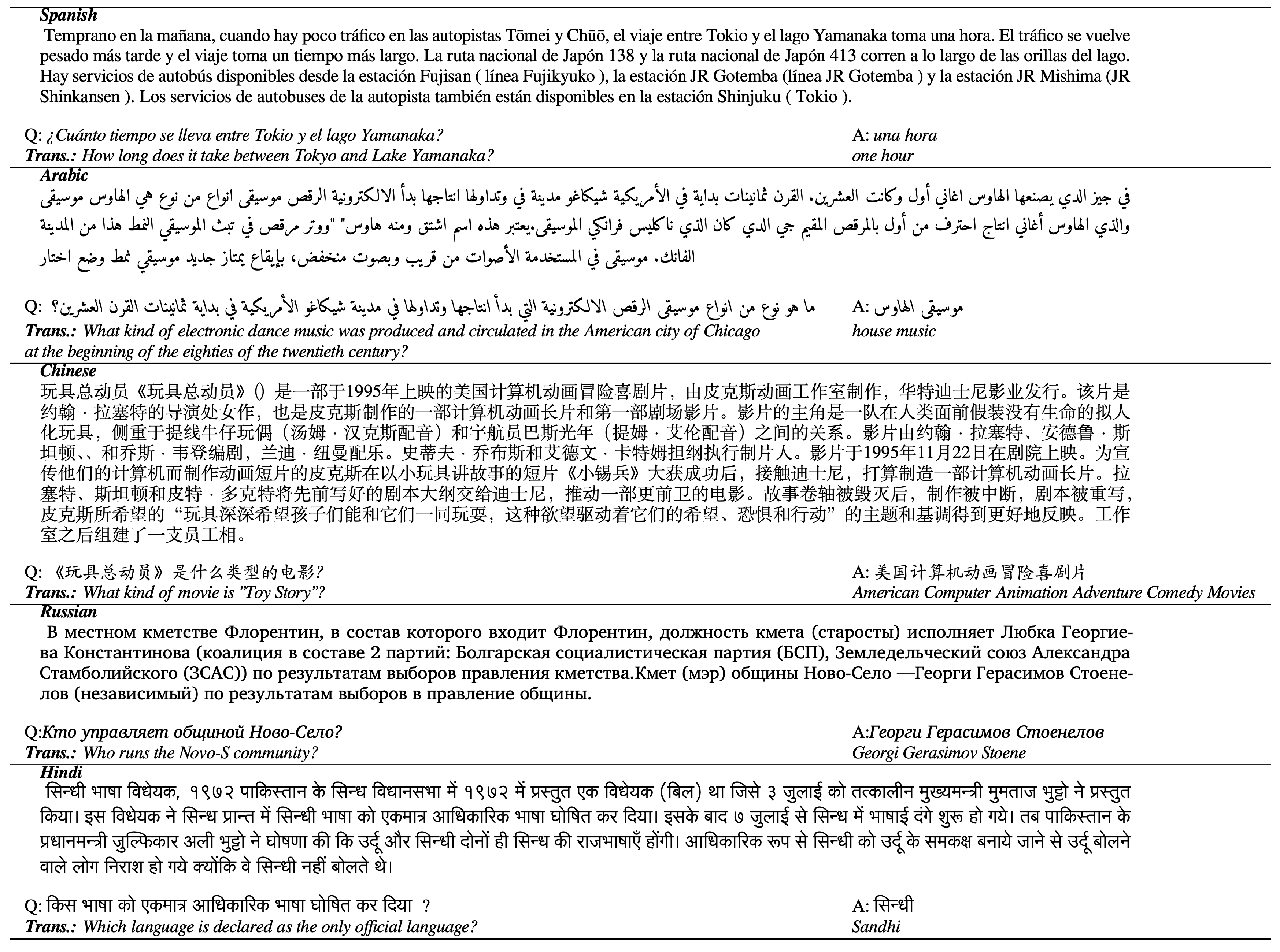}
    \caption{\small Samples of generated QAs in Spanish, Russian, Chinese, Arabic, and German. The generative model is mT5-XL fine-tuned on the mixture setting of section~\ref{sec:multi-task}. \textbf{\textit{Trans.}} refers to translations of the QA sample using Google Translate service.} 
  \label{fig:samples_multitask}
\end{figure*}

Fig.~\ref{fig:samples_multitask} demonstrates examples of generated samples in five languages using an mT5-XL (4B parameter) model fine-tuned in the multi-task setting (\ref{sec:multi-task}). It can be observed that: 1)~The generated questions are in the same language as the passage most of the time, 2)~The answers are relevant to the generated questions, 3)~The model is capable of generating long and non-trivial QA pairs.

\begin{figure*}[h]
    \centering
    \includegraphics[scale=0.28]{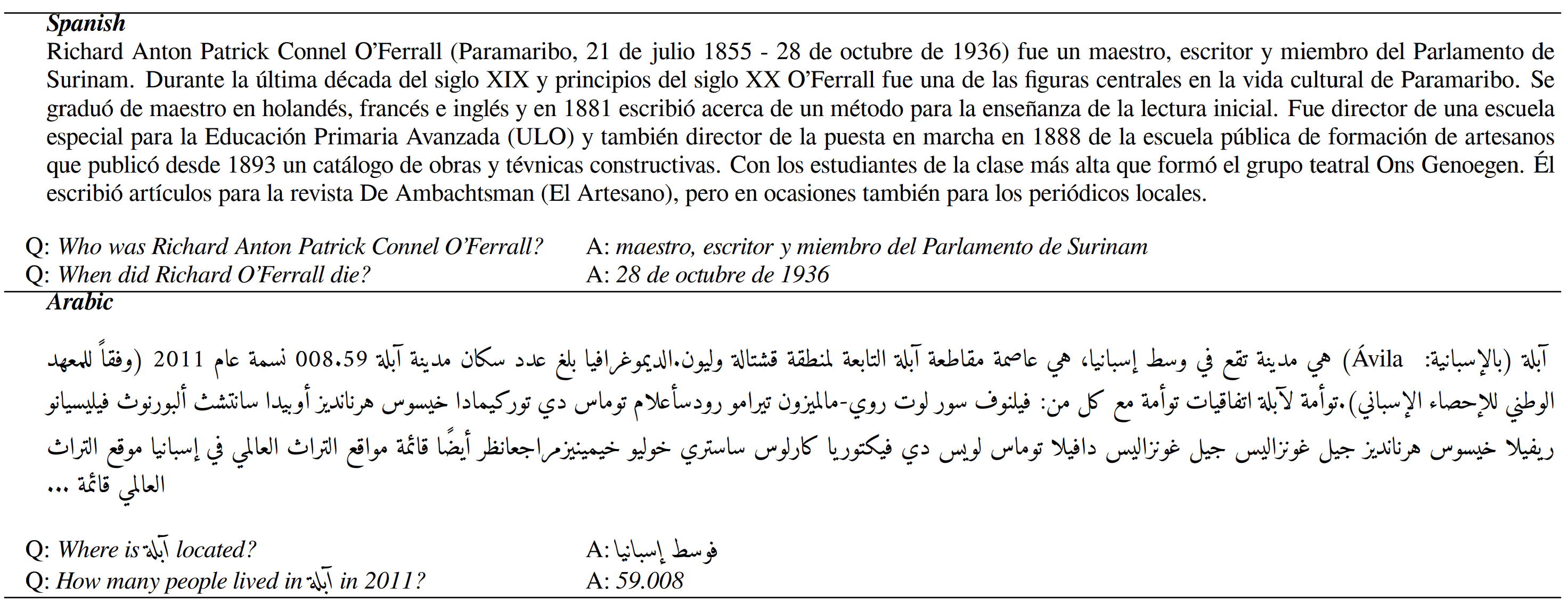}
    \caption{\small Samples of generated QAs in Spanish and Arabic. The mT5-XL model is unable to generate valid questions in the target language, as in this case it was fine-tuned exclusively on the English QA generation task from section~\ref{sec:qa_gen_task}.}
  \label{fig:samples_qa_gen}
\end{figure*}

Fig.~\ref{fig:samples_qa_gen} illustrates generated QA samples in Spanish and Arabic, when only the QA Generation task (\ref{sec:qa_gen_task}) is included in the fine-tuning. We observe: 1) Questions are primarily in English, not the target language, 2) Outputs contain certain tokens and entities mentioned in the language of the passage, 3) Ignoring language issues, the outputs exhibit semantically well-formed QA correspondence.

\subsection{Decoding and Filtering}\label{sec:filtering}



Since the accuracy of the generated question and answers is vital in improving the performance of downstream model, such generated samples require a strong filtering technique. Using the F1 score of a trained RC model to perform filtering, a.k.a.~\textit{round-trip} filtering, has been previously explored by \citet{puri2020training} and \citet{alberti-etal-2019-synthetic}. For a generated QA sample \textit{(q,~a,~p)}, where \textit{q}, \textit{a}, and \textit{p} indicate question, answer, and passage, the following steps are performed: 1) A trained RC model is applied to \textit{(q,~p)}, thus predicting \textit{$a'$}, and 2) The F1 score of \textit{a} and $a'$ is calculated, and if above a certain threshold, \textit{(q,~a,~p)} is kept, otherwise dropped.

%% file: related_work.tex
\section{Related Work}\label{sec:related}
Recent work has explored question-answer generation \cite{alberti-etal-2019-synthetic, puri2020training, lee2020generating,shakeri2020endtoend}, but limited in scope to English. We leverage the modeling and filtering approaches proposed by \citet{shakeri2020endtoend} due to their simplicity and effectiveness.

\citet{cross-ling-training} explores cross lingual question generation. In contrast to our work, this only generates questions, without the corresponding answers. Additionally, this approach requires a complicated pre-training process on the target languages, as well as gold samples to fine-tune the generative models, so it is not easily extensible to other languages. This is in contrast to our work, which does not require any gold QA samples in any language other than English. Another distinguishing factor is that we demonstrate improved performance on downstream QA tasks, while \citet{cross-ling-training} only measure the quality of the generated samples on automatic metrics such as BLEU, and human evaluations.

Similarly, \citet{chi2019crosslingual} explore cross-lingual question-only generation using SQuAD English samples. They propose cross-lingual pre-training on the source and target languages. Similar to \citet{cross-ling-training}, their focus is only on the quality of the generated questions, whereas we validate our approach directly through improvements on downstream QA tasks. Moreover, while \citet{chi2019crosslingual} depends on a complex pre-training recipe and parallel sentences in both source and target languages, our approach not only does not require such parallel corpus, but also the MLM task included in our fine-tuning setting is widely used and studied. This leads to our approach being more easily adaptable to other languages and pre-trained generative models.

Most closely related to our work is the multilingual synthetic question generation approach of \citet{riabi2020synthetic}. However, there are two important differences between the two approaches. Firstly, our work includes both question and answer generation using a single model, while theirs only focuses on question generation. We believe generating the question and answer jointly is a richer problem that better harnesses the capabilities of pre-trained language models. Their question generation is conditioned on the selected answers, which further limits the generation. Secondly, their proposed method depends on translated SQuAD to target languages to fine-tune the generative model, hence limiting the application of their approach to languages that such translation data does not exist. Even if such translated data exists in a target language, the quality of the samples generated by the generative model trained on such training data is highly affected by the quality of the translations. This could lead to low quality QA samples in low resource languages.  This is in contrast to our zero-shot approach, which does not require any training data in the target language.

%% file: experiments.tex
\section{Experimental Setup and Results}\label{sec:experiments}

\subsection{Datasets}
\textbf{SQuAD} \cite{squad} is an English QA dataset consisting of 100k samples. The passages are extracted from Wikiepdia. We use the train and dev splits of SQuAD 1.1 in this work. \\
\textbf{XQuAD} \cite{xquad} is a multilingual QA dataset consisting of 240 paragraphs and 1190 question-answers in Arabic, Chinese, German, Greek, Hindi, Russian, Spanish, Thai, Turkish and Vietnamese. These samples have been professionally translated from the SQuAD 1.1 dev set. \\
\textbf{MLQA} \cite{mlqa} is a benchmark dataset for evaluating cross-lingual question answering performance. This dataset contains over 5k QA instances (12k in English) following the SQuAD format in each of Arabic, Chinese, English, German, Hindi, Spanish and Vietnamese. We use the test split in our evaluations.\\ 
\textbf{TyDiQA} \cite{tydiqa} is another multilingual QA dataset consisting of 200k QA pairs from 11 typologically diverse languages. There is less lexical overlap between questions and answers compared to XQuAD and MLQA. We use the Gold Passage task, which includes \textasciitilde 50k samples in the train split and between 130 and 1100 samples for each language in the development set. 


\textbf{XTREME} \cite{xtreme} is a multilingual benchmark consisting of nine tasks spanning 40 typologically diverse languages. This dataset includes machine translated SQuAD 1.1 train and dev samples, which we employed in our experiments. We refer to such samples as \texttt{translate-train}.
\subsection{Generative Model Fine-tuning}\label{sec:genfine-tuning}
We used the official mT5-XL model \cite{mt5} with 3.7 billion parameters as the generative model. The official pre-trained checkpoint is fine-tuned using the mixture of tasks described in section~\ref{sec:modeling}. We chose the task mixing ratio to be 10:1, meaning for every 10 instances of the QA Generation task (\ref{sec:qa_gen_task}), we mix one instance of the MLM task (\ref{sec:mlm}). We experimented with other mixing ratios of 1000:1 and 100:1, as well, all of which under-performed 10:1.
The unsupervised MLM task covers text from two domains: 1) the subset of the mC4 corpus \cite{mt5} covering Arabic, Bengali, English, Finnish, Indonesian, Korean, Russian, Swahili, and Telugu, 2) questions from TyDiQA \cite{tydiqa} train and dev sets, covering the same set of languages. 

It is worth highlighting that we only fine-tune a single model to generate across all target languages. We do not apply language code prompts during fine-tuning or inference. We observe that by properly designing the fine-tuning mixture, the model is capable of generating samples in the language of the input passage. Human evaluations in section~\ref{sec:human-evals} further verify this.

All of our models are fine-tuned for 5000 steps with a batch size of 131,072 tokens, distributed over 64 TPU-v3 chips. The final checkpoint is used to perform synthetic data generation.


\subsection{Automatic Evaluation Results}

To compute automatic metrics such as BLEU against QA samples of the development set, we modify the generation task to generate a question given the passage and the answer. Conditioning on the answer is needed, as without it, the generative model might generate samples that are of high quality but not related to the answers provided in the development set for a given passage. This would lead to inaccuracies in interpreting metrics such as BLEU\footnote{All BLEU scores in this work are calculated using SacreBLEU v1.3.0 \cite{post-2018-call}, with ``exp'' smoothing and ``intl'' tokenization.}. 

\input{qgen}
\input{qgen_size}

Tab.~\ref{qgen} compares BLEU performance of two fine-tuning settings on the MLQA test set. We report results using the mT5-XL model. As can be seen, including the MLM tasks has a large impact on performance, conveying large gains up to +15 absolute BLEU points. This is in line with our observations from section~\ref{sec:filtering}, where adding MLM fine-tuning task enabled the generative model to produce QA samples in the language of the target passage. 

Interestingly, MLM on either mC4 or TyDiQA questions results in similar BLEU scores. Furthermore, adding a mixture of two does not yield any extra gains. However, eyeballing the generated samples indicated that the model fine-tuned on the mixture of both MLM tasks and the supervised English task generates more well-structured and sensible questions and answers. Human evaluations in section~\ref{sec:human-evals} verify the high quality of generated samples from a model trained with this mixture.

To investigate the effect of the generative model size on the quality of data generation, we perform experiments using  mT5 variants with different number of parameters: Base (600M), Large (1B) and XL (3.7B). We report results of the fine-tuned models with the mixture setting (\ref{sec:multi-task}) on the MLQA dataset in Tab.~\ref{qgen-size}. Model performance improves dramatically with the size of the pre-trained model. Based on these results, for the remainder of the paper, we use the mT5-XL model fine-tuned using the mixture approach.

\subsection{Human Evaluations}\label{sec:human-evals}

To perform manual quality evaluation of the generated questions, raters were presented with generated questions, and tasked with rating them according to the following criteria: 
\begin{itemize}[noitemsep]
    \itemsep0em 
    \item \textit{Is the question in the target language?} Raters could select \textit{yes} or \textit{no}.
    \item \textit{grammatical correctness:} Raters could select a whole number from 1 (lowest) to 4 (highest). 
    \item \textit{sensibility:} Raters could select a whole number from 1 (lowest) to 4 (highest). 
\end{itemize}

In total, 400 generated samples from 5 languages were randomly selected and rated by native speakers of each language. Each rater was assigned 40 samples. Two native speakers of each of the five languages were asked to perform the task. Tab.~\ref{tab:human_evals} shows the evaluation results. 

The results show that the multilingual generative model is nearly perfect at generating samples that match the language of the input passage. Considering no language codes are used during fine-tuning, and only the English supervised training data was used, the results show that our proposed mixture has enabled the model to perform zero-shot cross-lingual generation coherently.

Interestingly, Spanish samples achieve high scores in all of the categories. Considering the model is not trained on any Spanish samples, either in the MLM tasks or SQuAD~1.1, the model shows strong transfer learning capabilities. This implies that including the MLM task as proposed in our mixture setting not only prevents the generative model from \textit{catastrophic forgetting} of its multilingual capability on the languages included in the MLM fine-tuning task, but also on those not included.
The same argument partially applies to Hindi. While there are no Hindi samples in the fine-tuning mixture, Bengali (a related Indo-Aryan language) was seen in the MLM task.

\input{human_evals}

\section{Application of Synthetic Data to Multilingual Reading Comprehension}\label{sec:downstream_experiments}
In this section, we describe experimental results that demonstrate the efficacy of using synthetic samples for improving multilingual reading comprehension models. RC refers to the setting where given a passage and a question, the model is tasked with finding a span of the passage that answers the question.

\subsection{Synthetic Data Generation}\label{sec:synthgen}

We randomly selected 10k paragraphs from Wikipedia \cite{wikidump}, for each of Arabic~(ar), German~(de), Hindi~(hi), Russian~(ru) and Spanish~(es). The selected paragraphs were restricted to have between 30 and 450 tokens, thereby removing passages that are too long or too short.

We fine-tune the mT5-XL model according to the mixture setting discussed in section \ref{sec:multi-task} and the hyper-parameters from section \ref{sec:genfine-tuning}, and then use this model to generate 20 questions per passage. \textit{Top-k} sampling \cite{holtzman2020curious} with \textit{k=10} and temperature of 0.5 was applied. The generated samples are processed to ensure: 1) each consists of a question followed by an answer, 2) the answer does exist in the passage. This was done to ensure answers are extractive. Non-extractive or no-answer QA are outside the scope of this work.

As discussed in section \ref{sec:filtering}, round-trip filtering is applied to the generated QA samples. We used an mT5 XL model trained on SQuAD 1.1 \cite{squad11} as the filtering model. This process would result in approximately 10-20k synthetically generated samples per each target language. The generated samples are used in training the RC models.

\subsection{RC Model Fine-tuning}
All our reading comprehension models are initialized from the official mT5 \cite{mt5} and later fine-tuned on the generated samples. We experimented with Base (580M), Large (1.2B), and XL (3.7B) parameter variants of mT5. We fine-tune using the TensorFlow framework. Each model was trained for \textit{10000} steps with a learning rate of \textit{1e-3} and a batch size of \textit{131072} tokens. The models were trained on 16 TPU-v3 chips.
In experiments where both the SQuAD 1.1 samples and synthetically generated ones are used to fine-tune the RC models, the model is trained on a mixture of the two, with a mixing ratio of \textit{1}. 

\subsection{Results}
\input{xquad_mt5_base}
Tabs.~\ref{tab:xquad_mt5_base}--\ref{tab:xquad_mt5_xl} demonstrate the F1 performance of the RC models trained on SQuAD 1.1 samples as well as synthetic data generated as described in \ref{sec:synthgen} on mT5 Base, Large, and XL models. ``\textit{SQuAD en}'' refers to the original SQuAD~1.1 \cite{squad11} dataset in English. Our zero-shot baselines, denoted by \textit{(ours)} were slightly higher than those reported in \cite{mt5} \textit{(paper)}. 

\input{xquad_mt5_large}
We can observe that ~Augmenting \textit{SQuAD en} with synthetic samples leads to large gains with the base model. \textbf{+9} absolute point improvement can be observed with Russian. Furthermore, with the base model, all the average F1's are improved with the addition of synthetic data, regardless of which language the synthetic samples come from. The largest gain is seen when German samples are added \textbf{(+2.9)}.

\begin{figure}[h]
    \centering
    \includegraphics[scale=0.12]{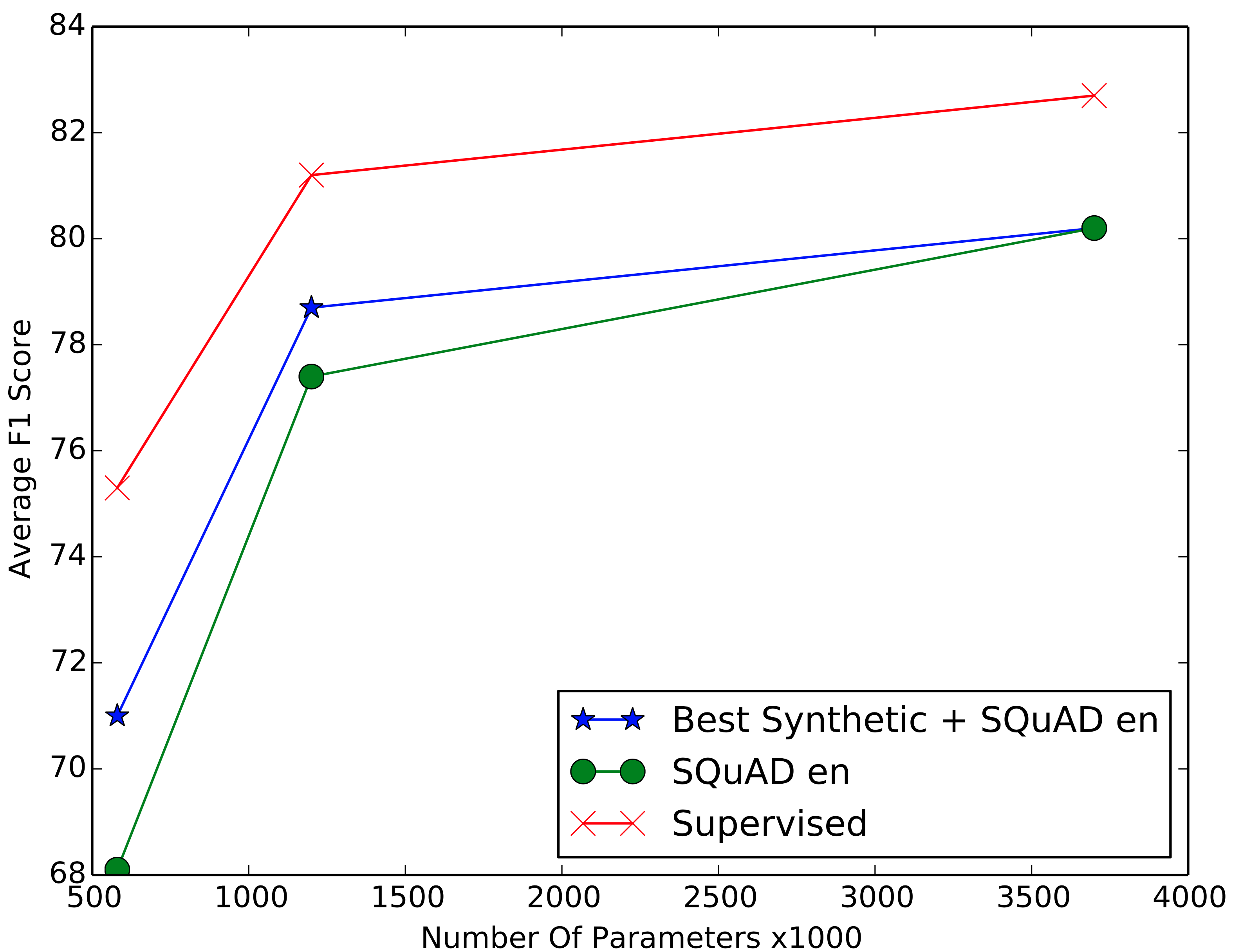}
    \caption{\small{Scaling effect on augmentation using synthetic samples.}}
  \label{fig:scaling_effect}
\end{figure}

As the number of parameters of the mT5 model increases, the gains through synthetic augmentation become less, as results show in Tabs.~\ref{tab:xquad_mt5_large} and \ref{tab:xquad_mt5_xl}. The maximum average F1 improvements of \textbf{+1.2} absolute points with the Large model is achieved. With the XL model, the average F1 scores are either the same as the zero-shot baseline or slightly lower. This is expected as the model size increases, the gap between zero-shot and supervised also becomes smaller, hence less headroom exists when adding the synthetic samples. Fig.~\ref{fig:scaling_effect} demonstrates this scaling effect. Nonetheless, improvements of \textbf{+5.1}, \textbf{+2.2}, and \textbf{+3.4} are observed on Russian, Arabic, and Greek, respectively with the mT5 Large model. Similarly, smaller per language gains can be seen with augmentation with the XL model, as shown in Tab.~\ref{tab:xquad_mt5_xl}.
\input{xquad_mt5_xl}


Comparing the \textit{Supervised} metrics vs.~\textit{SQuAD en + <lang>.}\ indicates that using synthetic samples reduces the gap between the zero-shot versus supervised performance of the trained RC models with Base and Large. This gap is reduced from \textbf{7.2} to \textbf{4.2} absolute points with the base model. However, there still exists a sizeable gap, which can be further close with using higher quality synthetic samples. 



%% file: qgen.tex
\begin{table}[]
\centering
\footnotesize
\scalebox{0.85}{
\begin{tabular}{l|rrrrrrr} \hline
Training Task          & ar  & de   & en   & es  & hi  & vi   & zh  \\ \hline
SQuAD en & 1.7 & 3.0  & 23.4 & 3.6 & 3.2 & 4.4  & 1.2 \\
Mixture 1 & 12.2&	14.9&	25.0 & 18.4&	10.6&	13.8&	10.0 \\ 
Mixture 2 & 13.1 & 15.2 & 24.9 & 18.4 & 11.1 & 13.9 & 9.7 \\ 
Mixture 3 & 14.5 & 14.8 & 25.0 & 18.6 & 10.8 & 13.5 & 9.6 \\ \hline
\end{tabular}
}
\caption{{\small Comparison of question generation quality (BLEU score) on the MLQA test set with mT5-XL: The Mixtures are as follows: \textit{SQuAD en}: SQuAD en as the training data, \textit{Mixture 1}: SQuAD en + MLM on mC4 subset, \textit{Mixture 2}: SQuAD en + TyDiQA questions, \textit{Mixture 3}: SQuAD en + MLM on mC4 subset + MLM on TyDiQA questions. }}
\label{qgen}
\end{table}

%% file: qgen_size.tex
\begin{table}[]
\centering
\footnotesize
\scalebox{0.86}{
\begin{tabular}{l|rrrrrrr} \hline
Model Size     & ar  & de   & en   & es  & hi  & vi   & zh  \\  \hline
Base (600M)  & 3.9 & 5.1  & 19.0 & 8.2 & 3.5 & 7.4  & 3.1 \\ 
Large (1B) & 10.3 & 5.7  & 23.9 & 5.9 & 4.3 & 6.2  & 3.9 \\
XL (4B)   & 14.5 & 14.8 & \textbf{25.0} & 18.6 & 10.8 & 13.5 & 9.6 \\
XXL (13B)   & \textbf{15.8} & \textbf{16.2} & \textbf{24.9} & \textbf{19.3} & \textbf{12.2} & \textbf{15.6} & \textbf{10.2} \\
\hline
\end{tabular}
}
\caption{{\small Performance of question generation (mixture setting) on the MLQA test set for different mT5 model sizes.}}
\label{qgen-size}
\end{table}

%% file: human_evals.tex

\begin{table}[]
\centering
\footnotesize
\scalebox{0.9}{
\begin{tabular}{l|rrr} \hline

& Target & Grammatical  &   \\ 
& Language & Correctness & Sensibility \\
\hline
Arabic & 0.98 & 3.55 & 3.35   \\
Chinese & 1.00 & 3.60 & 3.60  \\
Hindi & 1.00 & 2.93 & 3.35 \\
Russian & 1.00 & 3.50 & 3.75  \\
Spanish & 1.00 & 3.10 & 3.05  \\
\hline
Average & 1.00 & 3.34 & 3.38 \\
\hline
\end{tabular}
 }
\caption{{\small Human Evaluation metrics on the generated samples. Samples are randomly drawn, and rated by the native speakers of Arabic, Chinese, Hindi, Russian and Spanish.}}
\label{tab:human_evals}
\end{table}

%% file: xquad_mt5_base.tex
\begin{table*}[h!]
\centering
\footnotesize
\scalebox{0.95}{
\begin{tabular}{l|cccccccccccc}
\hline
\textbf{Dataset} & en & ar & de & el & es & hi & ru & th & tr & vi & zh & avg \\
\hline
SQuAD en (paper) & 84.6  & 63.8 & 73.8  & 59.6 & 74.8 & 60.3  & 57.8 & 57.6  & 67.9  & 70.7 & 66.1  & 67.0 \\ \hdashline
SQuAD en (ours) & \textbf{85.5} & 65.7 & 73.6 & 62.5 & 75.0 & 62.4& 61.9 & 57.6 & 68.9 & \textbf{71.9} & \textbf{71.1} & 68.1 \\
SQuAD en + ru & 83.7& 67.5 & 73.6 & 69.3 & 73.8 & \textbf{66.2} & 70.3 & 62.7 & 67.5 & 68.8 & 68.9 & 70.0 \\
SQuAD en + hi & 84.2 & 68.3 & 75.0 & 68.4 & 75.0 & 63.7 & 68.2 & \textbf{64.5} & 67.2 & 69.5 & 68.9 & 69.9 \\
SQuAD en + de & 84.6& \textbf{69.0} & 71.8 & \textbf{70.2} & \textbf{75.7} & \textbf{66.2} & \textbf{71.0} & 63.5 & \textbf{70.0} & 70.9& \textbf{71.2} & \textbf{71.0} \\
SQuAD en + ar & 84.5& 64.0& 74.4& 69.4& 74.4 & 65.1 & 65.1 & 62.5 & 67.9 & 70.0 & 70.2 & 70.2 \\
SQuAD en + es & 84.8 & \textbf{69.1} & \textbf{76.1} & 68.2 & 72.8 & 65.4 & 68.9 & 62.7 & 70.0 & 71.0 & 71.0 & 70.6 \\
\hline
Supervised & 83.1 & 72.4 & 76.9 & 76.8 & 79.0 & 71.4 & 76.1 & 67.9 & 72.5 & 75.9 & 76.9 & 75.3 \\
\hline
\end{tabular}
}
\caption{{\small Performance of fine-tuned mT5 Base models on XQuAD. \textit{Supervised} refers to training on SQuAD en + \texttt{translate-train} dataset of the target language.}}\label{tab:xquad_mt5_base}
\end{table*}

%% file: xquad_mt5_large.tex
\begin{table*}[h!]
\centering
\footnotesize
\scalebox{0.95}{
\begin{tabular}{l|cccccccccccc}
\hline
\textbf{Dataset} & en & ar & de & el & es & hi & ru & th & tr & vi & zh & avg \\
\hline
SQuAD en (paper) & 88.4 &  75.2 & 80.0 & 77.5 & 81.8 & 73.4 & 74.7 & 73.4 & 76.5 & 79.4 & 75.9 & 77.8 \\ \hdashline
SQuAD en (ours) & 88.6 & 75.0 & 80.4 & 76.5 & 81.6 & 73.9 & 74.1 & 73.8 & \textbf{76.2} & \textbf{80.1} & 76.4 & 77.4 \\
SQuAD en + ru & 88.2 & 76.6 & 81.2 & 79.1 & 82.6 & 76.1 & 77.6 & 72.1 & 75.1 & 78.4 & 77.4 & \textbf{78.6} \\
SQuAD en + hi & \textbf{88.9} & 76.7 & 81.5 & 79.4 & \textbf{82.9} & 73.4 & 78.7 & \textbf{74.1} & 75.0 & 79.1 & 78.0 & \textbf{78.6} \\
SQuAD en + de & 88.0 & 72.7 & 79.7 & 73.0 & 82.0 & 73.6 & 76.4 & 71.6 & 74.8 & 78.7 & 76.2 & 76.6 \\
SQuAD en + ar & 88.0 & 73.3 & 81.2 & 78.8 & 82.4 & 75.1 & 78.5 & 71.4 & 75.6 & 77.3 & \textbf{78.2} & 77.8 \\
SQuAD en + es & 88.2 & \textbf{77.2} & \textbf{81.8} & \textbf{79.9} & 81.3 & \textbf{76.4} & \textbf{79.2} & 72.3 & 75.8 & 79.5 & 77.7 & \textbf{78.7} \\
\hline
Supervised & 87.3 & 79.4 & 82.7 & 81.8 & 83.8 & 78.0 & 81.9 & 74.7 & 80.2 & 80.4 & 83.2 & 81.2 \\
\hline
\end{tabular}
}
\caption{{\small Performance of fine-tuned mT5 Large models on XQuAD. \textit{Supervised} refers to training on SQuAD en + \texttt{translate-train} dataset of the target language.}}\label{tab:xquad_mt5_large}
\end{table*}

%% file: xquad_mt5_xl.tex
\begin{table*}[h!]
\centering
\footnotesize
\scalebox{0.95}{
\begin{tabular}{l|cccccccccccc}
\hline
\textbf{Dataset} & en & ar & de & el & es & hi & ru & th & tr & vi & zh & avg \\
\hline
SQuAD en (paper) & 88.8 & 77.4 & 80.4 & 80.4 & 82.7 & 76.1 & 76.2 & 74.2 & 77.7 & 80.5 & 80.5 & 79.5  \\ \hdashline
SQuAD en (ours) & \textbf{89.7} & \textbf{79.2} & 80.9 & 80.9 & 83.2 & \textbf{78.7} & 78.4 & 74.3 & 78.4 & 79.5 & \textbf{80.7} & \textbf{80.2} \\
SQuAD en + ru & 89.1 & 78.6 & \textbf{82.1} & \textbf{81.7} & 82.7 & 78.6 & 79.4 & 74.3 & \textbf{78.7} & 80.6 & 79.2 & \textbf{80.2} \\
SQuAD en + hi & 89.1 & \textbf{79.1} & 81.7 & 80.9 & \textbf{83.4} & 76.1 & 79.0 & \textbf{74.6} & 77.6 & 81.0 & 80.4 & 79.9 \\
SQuAD en + de & 88.8 & 78.2 & 81.2 & \textbf{81.7} & 82.8 & 78.1 & \textbf{79.6} & 74.0 & 77.7 & \textbf{81.2} & 79.7 & 80.1 \\
SQuAD en + ar & 89.0 & 75.0 & 81.3 & 81.5 & 82.8 & 78.4 & 79.3 & 73.5 & 78.4 & 80.2 & 80.4 & 79.6 \\
SQuAD en + es & 88.8 & 79.0 & \textbf{82.2} & 81.3 & 82.6 & \textbf{78.7} & 78.8 & 73.8 & 78.3 & \textbf{81.1} & 80.5 & \textbf{80.2} \\
\hline
Supervised & 88.5 & 80.9 & 83.4 & 83.6 & 84.9 & 79.6 & 82.7 & 78.5 & 82.4 & 82.4 & 83.2 & 82.7  \\
\hline
\end{tabular}
}
\caption{{\small Performance of fine-tuned mT5 XL models on XQuAD. \textit{Supervised} refers to training on SQuAD en + \texttt{translate-train} dataset of the target language.}}\label{tab:xquad_mt5_xl}
\end{table*}

%% file: conclusion.tex
\section{Conclusion}\label{sec:conclusion}
In this work, we presented a simple yet effective approach to generate large-scale synthetic multilingual question-answer pair data, which can be used to improve the zero-shot performance of multilingual RC models. Our experimental results showed large improvements in the performance of RC models trained on our synthetic multilingual datasets as compared to standard zero-shot baselines. Moreover, our zero-shot generation approach proved to be easily applied to any language, as long as the language is supported by the pre-trained multilingual generative model.

While our results showed that using synthetic samples alongside English training data can significantly narrow the gap between zero-shot and supervised performance of RC models, the gap still remains. We are optimistic that future work can reduce this gap further through improved generation quality.

%% file: acknowledgement.tex
\section{Acknowledgement}\label{sec:acknowledgement}
We  thank Fadi Haik, Nizar Saqqar, Bahaa El-Taweal, Stanela Khalil, Igor Krivokon, Vladimir Magay, Tania Rojas-Esponda, and Gustavo Hernandez Abrego for  evaluations of the samples and providing valuable feedback on their quality.